\documentclass[10pt,twocolumn,letterpaper]{article}
\PassOptionsToPackage{table}{xcolor}
\usepackage{stfloats}
\usepackage[pagenumbers]{cvpr}      %

\usepackage{microtype}

\renewcommand{\paragraph}[1]{\vspace{.5em}\noindent\textbf{#1.}}

\definecolor{cvprblue}{rgb}{0.21,0.49,0.74}
\usepackage[pagebackref,breaklinks,colorlinks,allcolors=cvprblue]{hyperref}

\usepackage{booktabs}
\usepackage{multirow}
\usepackage{makecell}
\usepackage{pifont}
\usepackage{graphicx}
\usepackage{float}
\usepackage{needspace}
\usepackage{array}
\usepackage{tabularx}
\usepackage{placeins}

\newcommand{\cmark}{\ding{51}}
\newcommand{\xmark}{\ding{55}}
\newcommand{\ours}{\textsc{AGPF}}
\newcommand{\farrope}{FA-RoPE}
\newcommand{\apbox}{AP$^{\rm box}$}
\newcommand{\apmask}{AP$^{\rm mask}$}

\newcommand{\methodpair}[2]{%
    \multirow{2}{*}{%
        \makecell[l]{%
            #1\\[-1.0pt]
            {\scriptsize\itshape #2}%
        }%
    }%
}

\newcommand{\aghl}[1]{%
    \cellcolor{black!6}#1%
}

\newcommand{\aghlb}[1]{%
    \cellcolor{black!6}\textbf{#1}%
}

\newcommand{\apgain}[2]{%
    \textbf{#1}\kern0.12em%
    {\color{black!65}\tiny\raisebox{-0.32ex}{+#2}}
}

\newcommand{\aghlgain}[2]{%
    \cellcolor{black!6}\apgain{#1}{#2}%
}

\def\confName{CVPR}
\def\confYear{2027}

\title{AGPF: Dynamic Source Selection for Feature Pyramid Fusion}

\author{
X.~Yang\qquad G.~Agam\\
Department of Computer Science\\
Illinois Institute of Technology\\
{\tt\small xyang99@hawk.illinoistech.edu\quad agam@illinoistech.edu}
}

\hypersetup{
    pdftitle={AGPF: Dynamic Source Selection for Feature Pyramid Fusion},
    pdfauthor={Xu Yang and Gady Agam}
}

\begin{document}
\maketitle

\begin{abstract}
    Feature pyramids provide dense predictors with features at several spatial resolutions.
    However, it remains largely unexplored which pyramid levels each output location should aggregate information from.
    We investigate this question by allowing each output location to attend to features from every pyramid level and measuring how attention is distributed across scales.
    Surprisingly, locations on the finest output grid allocate approximately 75\% of their attention to features two levels coarser,
    rather than to the current or adjacent levels.
    Even on the same output grid, the balance shifts toward coarser features as object size increases.

    These observations motivate Asymmetric Global Pyramid Fusion (AGPF),
    a neck that preserves the original output grids while allowing each location to dynamically select its contextual sources across scales.
    Each location interacts with a multiscale memory whose features are aligned in image coordinates,
    enabling flexible cross-scale aggregation without altering prediction resolutions.
    Compared with FPN, AGPF improves all 19 object detection and all seven instance segmentation configurations evaluated on COCO.
    A single fusion block improves Cascade Mask R-CNN by 1.7 box AP and 2.3 mask AP, and the gains consistently transfer to semantic and panoptic segmentation.
\end{abstract}

\begin{figure}[t]
    \centering
    \includegraphics[width=\columnwidth]{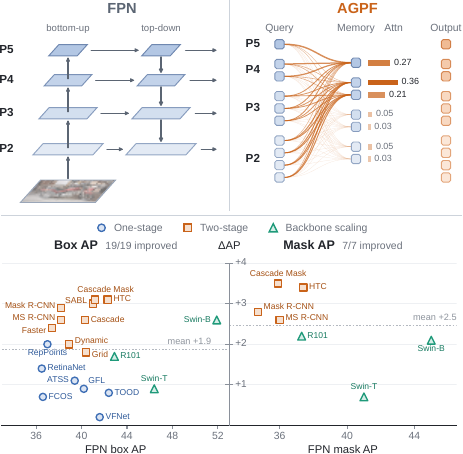}
    \caption{
        \textbf{From fixed fusion paths to per-location source selection.}
        \emph{Top:} FPN adds lateral and top-down connections to the
        bottom-up backbone; AGPF lets each output location retrieve from a
        shared multilevel memory, normalizing attention jointly over all
        levels.
        \emph{Bottom:} box and mask AP gains over matched FPN baselines,
        one point per configuration; dashed lines mark the means.
    }
    \label{fig:agpf_generality}
    \vspace{-1em}
\end{figure}

\begin{figure*}[h]
    \vspace{-0.5em}
    \centering
    \makebox[\textwidth][c]{
        \includegraphics[width=1.04\textwidth]
        {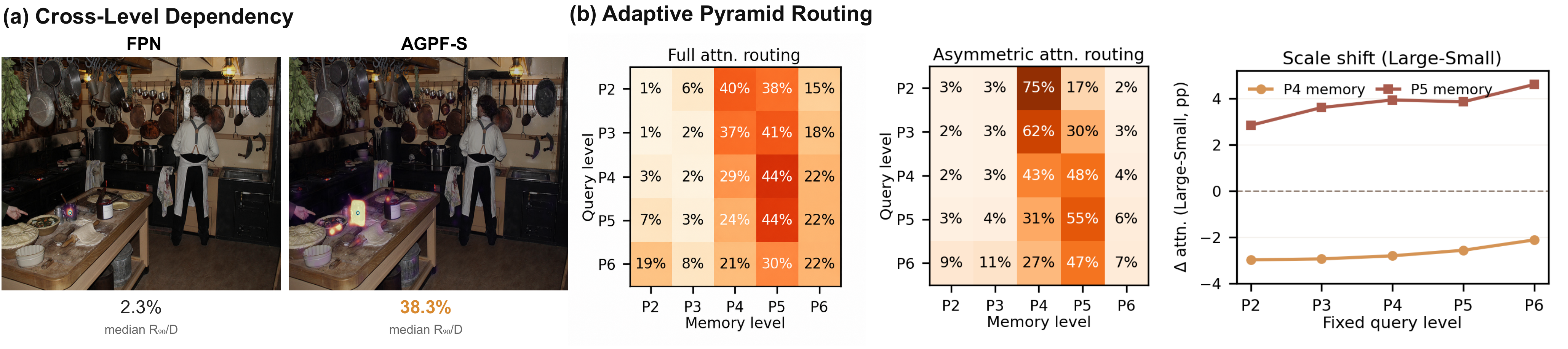}
    }
    \caption{
        \textbf{Pyramid source selection and spatial dependency.}
        \textbf{(a)} Gradient-based dependency maps for a matched query; the
        normalized $90\%$-mass radii are medians over $600$ paired queries.
        \textbf{(b)} Source-level attention mass with full and pooled memory,
        averaged over object-centered queries and heads; the right plot shows
        the large-minus-small change in pooled-memory routing at each fixed
        query level.
    }
    \label{fig:analysis}
    \vspace{-1em}
\end{figure*}

\section{Introduction}
Feature pyramids provide representations of an image at multiple spatial resolutions.
Dense predictors rely on these representations for complementary localization detail and semantic context, while a neck determines how information is exchanged among them.
This organization naturally invites a structural view of fusion: connect pyramid levels so that their features can support one another.
Yet the level at which a feature is needed for prediction does not determine where its supporting context should come from.
A fundamental question is therefore which pyramid features each output location should combine.

Existing necks approach this question through different interaction structures.
FPN~\cite{lin2017feature} propagates coarse features along a top-down pathway, while subsequent designs add bottom-up aggregation, search for cross-scale connections, or repeat bidirectional fusion~\cite{liu2018path,ghiasi2019fpn,tan2020efficientdet}.
Attention-based necks and multi-scale encoders introduce content-dependent interactions, organizing context through directional cross-level transformations~\cite{zhang2020feature}, shared global features~\cite{gu2022content}, or grouped cross-layer attention~\cite{du2025cross}.
Other designs fuse pyramid features resampled to common spatial resolutions~\cite{dai2021dynamic,wang2023gold}, retrieve from query-dependent sampling points with a fixed budget per level~\cite{zhu2021iclr-deformable}, or combine self-attention at the coarsest encoder level with convolutional cross-scale fusion~\cite{zhao2024detrs}.
Despite these diverse designs, they all specify how pyramid levels interact rather than asking which levels each output location actually depends on.
Consider two locations on the finest output grid, one on a small object and the other on a large one.
In FPN, both receive coarse context through the adjacent level, where features from farther up the pyramid have already been combined.
Adding a cross-level shortcut gives both locations direct access to a more distant source.
If each location could draw directly from every level, would nearby levels contribute most, and would the two locations favor the same sources?

We let each output location attend to a shared memory of pooled features from all pyramid levels and measure the share of attention it assigns to each level (Fig.~\ref{fig:analysis}(b)).
Over object-centered queries on COCO, attention concentrates on P4 and P5.
Queries on the finest level, P2, allocate approximately 75\% of their attention to P4, two levels away; this preference persists after accounting for the number of memory tokens at each level.
The concentration does not imply a fixed assignment: even with the query level held constant, attention shifts from P4 toward P5 as object size increases.
Thus, the finest output grid can draw most of its retrieved context from a nonadjacent source, while the balance among sources changes with the represented object.

These findings suggest organizing pyramid fusion around the sources selected by each location (Fig.~\ref{fig:agpf_generality}, top).
The output hierarchy specifies the resolutions at which features are produced; the observed retrieval patterns describe a different organization of how context is used.
We develop this view into Asymmetric Global Pyramid Fusion (AGPF), a neck that gives each output location direct access to a shared multilevel spatial memory and learns the contributions of its sources.
This formulation preserves the native output grids while allowing each query to compare context across both levels and spatial regions.
Complementary gradient-based analysis shows that AGPF retains strong local sensitivity while exhibiting a broader spatial dependence on the input pyramid than FPN (Fig.~\ref{fig:analysis}(a)).

Joint retrieval across resolutions also requires a consistent spatial reference.
A displacement of one cell represents different image-space distances at different pyramid levels, and memory pooling changes those distances further.
We introduce Feature-Aligned Rotary Position Embedding (FA-RoPE), which places each query and memory token at the image-space center of its sampling cell before applying rotary encoding.
Cross-level interactions therefore use a common spatial reference, accounting for both backbone stride and memory pooling.

Making source selection practical also requires controlling the size of the context.
The two finest levels account for over 90\% of the tokens in an uncompressed pyramid, but preserving their dense output grids need not require equally dense representations on the memory side.
We therefore retain every output query and spatially pool the fine-level memory, keeping sources from all levels directly accessible.
A controlled comparison shows that this compact memory matches full-pyramid attention in box and mask AP with far fewer query--key interactions, so compression need not restrict retrieval to a subset of pyramid levels.

AGPF integrates source selection, spatial alignment, and memory pooling into a compact asymmetric attention block whose output retains the multiresolution interface expected by existing prediction heads.
Stacking this block without changing that interface yields AGPF-S, AGPF-B, and AGPF-L.

Replacing FPN with AGPF improves all 19 evaluated object-detection and all seven instance-segmentation configurations on COCO,
spanning one-stage, two-stage, and cascade detectors with convolutional and Transformer backbones (Fig.~\ref{fig:agpf_generality}, bottom),
and the gains extend to semantic segmentation on Cityscapes and ADE20K and to panoptic segmentation.
Together, the routing analysis and the task results support a view of pyramid fusion centered on per-location source selection:
dense multiresolution outputs can be served by selective retrieval from a compact, spatially aligned context.

\begin{figure*}[t]
    \vspace{-0.5em}
    \centering
    \includegraphics[width=\textwidth]{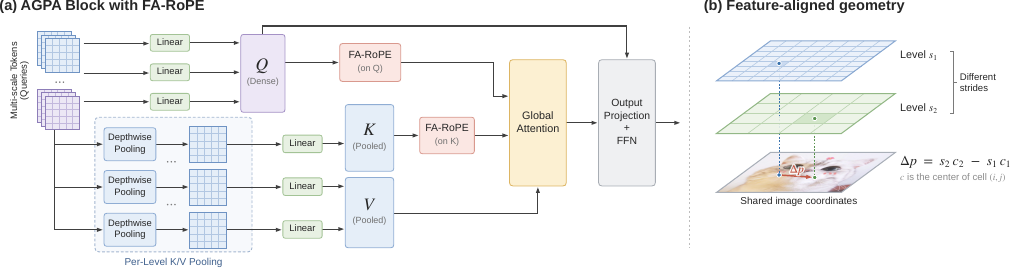}
    \caption{
        \textbf{AGPF block and feature-aligned coordinates.}
        \textbf{(a)} Dense queries jointly retrieve from pooled memory
        spanning all pyramid levels, preserving the output grids.
        \textbf{(b)} Shared image-space coordinates account for differing
        effective strides of queries and memory tokens.
    }
    \label{fig:agpf_framework}
    \vspace{-1em}
\end{figure*}

\section{Related Work}
\paragraph{Feature-pyramid fusion}
Existing methods differ in both cross-level connectivity and the way multiscale context is aggregated.
FPN combines top-down semantic enrichment with lateral connections~\cite{lin2017feature}.
Subsequent designs introduce bottom-up path augmentation~\cite{liu2018path},
search over cross-scale connections~\cite{ghiasi2019fpn},
and repeated bidirectional weighted fusion~\cite{tan2020efficientdet}.
Related designs further address how context is assembled: Dynamic Head combines scale-, spatial-, and task-aware attention~\cite{dai2021dynamic},
while Gold-YOLO gathers multilevel features and distributes the fused information back to individual levels~\cite{wang2023gold}.

\paragraph{Cross-level attention}
Attention-based pyramid modules provide direct precedents for content-dependent cross-scale interaction.
FPT uses separate self-level, top-down, and bottom-up transformations~\cite{zhang2020feature}.
CA-FPN augments pyramid features by attending to a shared global content map derived from the deepest backbone stage~\cite{gu2022content}.
CFPT combines grouped cross-layer channel and spatial attention with consistent positional encoding~\cite{du2025cross}.
In AGPF, pooled spatial tokens from all levels form a globally shared memory, allowing each native-resolution query to allocate attention jointly across levels and positions.
We use this formulation to examine how source allocation relates to output level and object scale.

\paragraph{Efficient multiscale attention}
Multiscale models reduce attention cost through sparse sampling, selective feature updates, and pooled representations.
Deformable DETR retrieves from a small set of query-dependent locations at each level~\cite{zhu2021iclr-deformable}.
Lite DETR interleaves updates across levels to reduce high-resolution processing~\cite{li2023lite},
while RT-DETR combines self-attention at the coarsest feature level with convolutional cross-scale fusion~\cite{zhao2024detrs}.
MViT and MViTv2 use pooling attention to construct efficient hierarchical backbone representations~\cite{fan2021multiscale, li2022mvitv2}.
AGPF applies memory compression to an existing pyramid while preserving the dense query grids used by prediction heads, allowing a compact multilevel context to serve output locations across resolutions.

\section{Approach}
\label{sec:approach}
AGPF treats the feature at each output location as a query
that retrieves context from a shared multilevel memory.
Each block preserves dense queries, pools the keys and values,
and aligns their spatial coordinates before attention
(Fig.~\ref{fig:agpf_framework}(a)).
The resulting features retain the original pyramid
resolutions for the prediction heads.

\paragraph{Pyramid tokenization}
Given an image, we form the input pyramid
$\{C_l\}_{l\in\mathcal{L}}$ from backbone features, using the levels
required by the prediction architecture.
Level $l$ has spatial size $H_l\times W_l$ and image stride $s_l$.
We project each level to a common width $d$ and flatten it into
$X_l\in\mathbb{R}^{N_l\times d}$, where $N_l=H_lW_l$.
The full token sequence is
$X=\operatorname{Concat}_{l\in\mathcal{L}}X_l$,
with each token associated with its source level and spatial location.

\paragraph{Memory pooling}
Within each block, we add a learned level embedding to $X_l$ and
use linear projections shared across levels to obtain queries $Q_l$,
keys $K_l$, and values $V_l$.
Queries retain every spatial location.
Keys and values are pooled separately on their spatial grids and
then concatenated across levels:
\begin{equation}
    Q=\operatorname{Concat}_{l}Q_l,\quad
    [K,V]=\operatorname{Concat}_{l}\mathcal{P}_l([K_l,V_l]).
    \label{eq:pyramid_memory}
\end{equation}
Here, $\mathcal{P}_l$ applies spatial pooling at stride $r_l$
independently to the key and value branches.
We use depthwise convolution for pooling, with separate parameters
for the two branches.
Levels with $r_l=1$ skip pooling.
Our default configuration compresses the finer levels, which contain
most of the pyramid tokens, while retaining the coarser levels at
their native resolutions.
Tokens from different levels remain separate in the memory, even
when pooling gives them the same spatial resolution.

\paragraph{Aligned spatial reference}
Queries and memory tokens can have different strides, so the same
grid index can refer to different image locations.
Feature-Aligned Rotary Position Embedding (FA-RoPE) expresses
their positions in a common image coordinate system
(Fig.~\ref{fig:agpf_framework}(c)).
Common-coordinate rotary embeddings have been used for
multi-resolution crops in MuViT~\cite{Mantes_2026_CVPR} and different
patch sizes in HyperDiT~\cite{he2026hyperdit}.
FA-RoPE applies this principle to native pyramid queries and
independently pooled memory, accounting for both the backbone stride
and the pooling stride.
We place each token at the image-space center of its sampling cell:
\begin{equation}
    p(i,j;t)=
    \left(\left(j+\tfrac12\right)t,\,
    \left(i+\tfrac12\right)t\right),
    \label{eq:farope_coordinates}
\end{equation}
where $(i,j)$ is the row--column index and $t$ is the effective stride.
Queries at level $l$ use $t=s_l$, while pooled keys use $t=r_ls_l$.
We apply mixed RoPE~\cite{heo2024rotary} to queries and keys using
these coordinates, with the rotary mapping shared across levels.
Their positional interactions therefore encode displacement in
common image units across resolutions.

\begin{figure}[t]
    \centering
    \includegraphics[width=\linewidth]{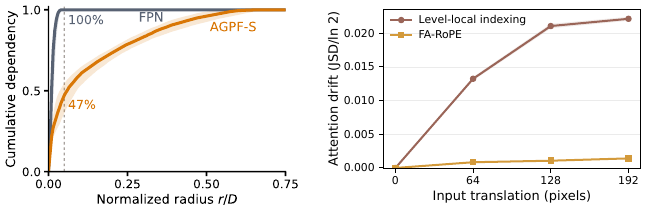}
    \caption{
        \textbf{Spatial dependency and translation consistency.}
        \emph{Left:} median cumulative gradient mass around the query
        (shading: interquartile range; dashed line: $r/D=0.05$).
        \emph{Right:} mean attention drift under input translation.
    }
    \label{fig:positional_consistency}
    \vspace{-1em}
\end{figure}

\paragraph{Joint pyramid retrieval}
We implement joint retrieval as asymmetric multi-head attention
between dense queries and pooled keys and values.
Each query attends to every memory token, with the softmax normalized
jointly over all source levels and spatial positions within each head.
Even queries on the same output grid can therefore assign different
weights to source levels and spatial regions.
Before output projection, we add the unrotated projected query to
the retrieved context.
A standard FFN follows the attention update; both sublayers use
residual connections and post-normalization.

Let $N_q=\sum_l N_l$ and $N_m=\sum_l\bar N_l$ denote the numbers
of query and memory tokens, where $\bar N_l$ is the token count
after pooling level $l$.
The attention computation is reduced from
$\mathcal{O}(N_q^2d)$ for full-pyramid attention to
\begin{equation}
    \mathcal{O}(N_qN_md), \qquad N_m\ll N_q,
    \label{eq:attention_complexity}
\end{equation}
while preserving an output at every query location.
Setting $r_l=1$ for all levels recovers the full-memory reference
with $N_m=N_q$.

\paragraph{Simple implementation and stacking}
A single standard attention operation handles all source levels,
without separate fusion modules for individual level pairs.
After each block, we split the output sequence by query level and
reshape it into feature maps $Y_l$ on the original $H_l\times W_l$
grids, preserving the multiresolution interface.
Blocks can therefore be stacked without modifying the prediction
heads, each constructing its memory from the updated features;
varying their number yields AGPF-S, AGPF-B, and AGPF-L.

\paragraph{Source selection}
We measure source selection using the attention mass assigned to
each pyramid level.
To account for different numbers of memory tokens across levels,
we also report enrichment relative to uniform attention over all
memory tokens.
For a query $q$ at level $l$, these measures are
    {\small
        \begin{equation}
            \begin{aligned}
                R_{l\rightarrow m}(q)
                 & = \frac{1}{n_h}\sum_{h=1}^{n_h}
                \sum_{j\in\mathcal{M}_m} A^{(h)}_{qj},                  \\
                E_{l\rightarrow m}(q)
                 & = \frac{R_{l\rightarrow m}(q)}{|\mathcal{M}_m|/N_m},
            \end{aligned}
            \label{eq:routing_mass}
        \end{equation}
    }
where $A^{(h)}_{qj}$ is the attention weight on memory token $j$
in head $h$, $\mathcal{M}_m$ contains the tokens from source
level $m$, and $n_h$ is the number of heads.
Uniform attention over all memory tokens gives $E=1$;
values above one indicate more attention to that source
than this reference assigns.

\paragraph{Spatial dependency}
We measure the sensitivity of an output feature to the input pyramid.
For a query $q$ at output level $l$, let $z_q=\|Y_l(:,q)\|_2$.
Its sensitivity to location $(x,y)$ in input level $m$ is
    {\small
        \begin{equation}
            g_{q,m}(x,y)
            =
            \left\|
            \frac{\partial z_q}{\partial C_m(:,x,y)}
            \right\|_2.
            \label{eq:dependency}
        \end{equation}
    }
We align these maps in image space, sum them on a common
reference grid, and normalize the result to unit mass.
We report $R_{90}/D$, where $R_{90}$ is the smallest query-centered
radius containing $90\%$ of this mass and $D$ is the reference-grid
diagonal.

\paragraph{Translation consistency}
We translate the input image, recompute its features, and compare
attention at corresponding spatial locations.
For each query and source level, let $\mathbf{a}$ and
$\mathbf{a}^{\Delta}$ denote the aligned attention distributions
before and after a translation by $\Delta$.
Both distributions are normalized within that source level.
We measure attention drift using normalized Jensen--Shannon
divergence,
$D_{\mathrm{attn}}
    =\operatorname{JSD}(\mathbf{a},\mathbf{a}^{\Delta})/\ln 2$.
Lower values indicate less attention drift.

\begin{table*}[t]
    \vspace{-0.5em}
    \centering
    \scriptsize
    \renewcommand{\arraystretch}{0.96}
    \setlength{\tabcolsep}{2.60pt}
    \resizebox{\linewidth}{!}{
        \begin{tabular}{lllccccccccccccc}
            \toprule[1.19pt]
            \multirow{2}{*}{Detector}     & \multirow{2}{*}{Neck / Fusion}   & \multirow{2}{*}{Source} & \multicolumn{6}{c}{Box Detection} & \multicolumn{6}{c}{Instance Segmentation} & \multirow{2}{*}{Mem.}                                                                                                                                                        \\
            \cmidrule[0.41pt](lr){4-9}
            \cmidrule[0.41pt](lr){10-15}
                                          &                                  &                         & AP                                & AP$_{50}$                                 & AP$_{75}$             & AP$_S$        & AP$_M$        & AP$_L$        & AP            & AP$_{50}$     & AP$_{75}$     & AP$_S$        & AP$_M$        & AP$_L$        &      \\
            \midrule[0.60pt]

            \multirow{6}{*}{Faster R-CNN} & FPN~\cite{lin2017feature}        & Reimpl.                 & 37.4                              & 58.1                                      & 40.4                  & 21.2          & 41.0          & 48.1          & --            & --            & --            & --            & --            & --            & 3777 \\
                                          & BiFPN~\cite{tan2020efficientdet} & Reimpl.                 & 35.3                              & 54.9                                      & 38.1                  & 20.3          & 38.7          & 45.3          & --            & --            & --            & --            & --            & --            & 5540 \\
                                          & HRFPN~\cite{sun2019high}         & Reimpl.                 & 37.4                              & 58.3                                      & 40.5                  & 21.4          & 40.9          & 48.3          & --            & --            & --            & --            & --            & --            & 6632 \\
                                          & PAFPN~\cite{liu2018path}         & Reimpl.                 & 37.7                              & 58.5                                      & 41.0                  & 22.0          & 41.4          & 48.7          & --            & --            & --            & --            & --            & --            & 3631 \\
                                          & AugFPN~\cite{guo2020augfpn}      & Reimpl.                 & 38.1                              & 59.5                                      & 41.4                  & 22.9          & 41.6          & 49.1          & --            & --            & --            & --            & --            & --            & 3583 \\
                                          & {\bf \ours-S}                    & Ours                    & \textbf{38.8}                     & \textbf{60.1}                             & \textbf{41.7}         & \textbf{23.0} & \textbf{41.9} & \textbf{51.0} & --            & --            & --            & --            & --            & --            & 4988 \\

            \midrule[0.47pt]
            \multirow{6}{*}{Mask R-CNN}
                                          & FPN~\cite{lin2017feature}        & Reimpl.                 & 38.2                              & 58.8                                      & 41.4                  & 21.9          & 40.9          & 49.5          & 34.7          & 55.7          & 37.2          & \textbf{18.3} & 37.4          & 47.2          & 4441 \\
                                          & BiFPN~\cite{tan2020efficientdet} & Reimpl.                 & 36.3                              & 55.8                                      & 39.5                  & 20.7          & 39.6          & 47.1          & 32.5          & 52.6          & 34.5          & 15.1          & 35.0          & 47.3          & 6240 \\
                                          & HRFPN~\cite{sun2019high}         & Reimpl.                 & 38.0                              & 58.6                                      & 41.4                  & 21.9          & 41.3          & 49.8          & 34.3          & 55.4          & 36.4          & 16.0          & 36.8          & 50.2          & 6675 \\
                                          & PAFPN~\cite{liu2018path}         & Reimpl.                 & 38.4                              & 59.1                                      & 41.7                  & 22.3          & 41.5          & 49.8          & 34.9          & 55.8          & 37.0          & 16.2          & 37.1          & 50.8          & 4323 \\
                                          & AugFPN~\cite{guo2020augfpn}      & Reimpl.                 & 39.1                              & \textbf{60.1}                             & \textbf{42.6}         & \textbf{23.1} & \textbf{42.7} & 50.5          & 35.2          & 56.8          & 37.5          & 16.7          & 38.0          & 51.0          & 4315 \\
                                          & {\bf \ours-S}                    & Ours                    & \textbf{39.2}                     & \textbf{60.1}                             & 42.5                  & 23.0          & 42.2          & \textbf{51.9} & \textbf{36.0} & \textbf{57.0} & \textbf{38.5} & 17.4          & \textbf{38.6} & \textbf{52.5} & 5260 \\

            \midrule[0.47pt]
            \multirow{7}{*}{Cascade Mask R-CNN}
                                          & FPN~\cite{lin2017feature}        & Reimpl.                 & 41.2                              & 59.4                                      & 45.0                  & 23.9          & 44.2          & 54.4          & 36.0          & 56.6          & 38.4          & \textbf{19.4} & 38.5          & 49.3          & 5970 \\
                                          & BiFPN~\cite{tan2020efficientdet} & Reimpl.                 & 38.1                              & 55.3                                      & 41.6                  & 20.3          & 41.3          & 50.8          & 33.1          & 52.4          & 35.5          & 14.6          & 35.5          & 48.6          & 7684 \\
                                          & HRFPN~\cite{sun2019high}         & Reimpl.                 & 40.9                              & 59.2                                      & 44.6                  & 23.1          & 44.2          & 54.0          & 35.7          & 56.2          & 38.3          & 16.8          & 38.2          & 52.5          & 7484 \\
                                          & PAFPN~\cite{liu2018path}         & Reimpl.                 & 41.2                              & 59.2                                      & 45.1                  & 23.5          & 44.8          & 54.6          & 35.9          & 56.3          & 38.5          & 16.5          & 38.6          & 52.7          & 5797 \\
                                          & AFPN~\cite{yang2023afpn}         & Reimpl.                 & 41.8                              & 60.5                                      & 45.6                  & 24.4          & 45.3          & 55.2          & 36.4          & 57.3          & 38.9          & 17.5          & 38.9          & 53.4          & 5989 \\
                                          & AugFPN~\cite{guo2020augfpn}      & Reimpl.                 & 42.1                              & 60.9                                      & 45.9                  & 24.8          & 45.6          & 55.4          & 36.5          & 57.8          & 39.2          & 17.9          & 39.1          & 53.3          & 5774 \\
                                          & FPT$^{\ast}$~\cite{zhang2020feature}      & Reimpl.                 & 42.1                              & 60.7                                      & 45.6                  & 23.3          & 45.8          & 55.6          & 36.5          & 57.5          & 39.0          & 17.0          & 38.5          & 52.8          & 9003 \\
                                          & {\bf \ours-S}                    & Ours                    & \textbf{42.9}                     & \textbf{61.7}                             & \textbf{46.8}         & \textbf{25.6} & \textbf{45.9} & \textbf{56.9} & \textbf{38.3} & \textbf{59.3} & \textbf{41.3} & 19.0          & \textbf{40.7} & \textbf{55.6} & 6389 \\

            \bottomrule[1.19pt]
        \end{tabular}
    }
    \caption{
        Comparison with feature-pyramid necks on COCO \texttt{val2017}
        under matched detector settings.
        AGPF-S uses one fusion block.
        Mem. denotes training memory in MB. $^{\ast}$ FPT builds on FPN; removing it reduces AP by over 2 points, indicating not fair comparison.
    }
    \label{tab:neck_detector_comparison}
    \vspace{-1em}
\end{table*}

\begin{table}[t]
    \vspace{-0.5em}
    \centering
    \setlength{\tabcolsep}{1.30pt}
    \renewcommand{\arraystretch}{0.98}
    \footnotesize
    \resizebox{\columnwidth}{!}{%
        \begin{tabular}{@{}l l lll lll@{}}
            \toprule[1.35pt]
            Method
             & Fusion
             & \apbox
             & AP$^{\rm box}_{50}$
             & AP$^{\rm box}_{S}$
             & \apmask
             & AP$^{\rm mask}_{50}$
             & AP$^{\rm mask}_{S}$                              \\
            \midrule[0.70pt]

            \multicolumn{8}{@{}l}{
                \textit{One-stage detection (ResNet-50)}
            }                                                   \\

            \methodpair{RetinaNet~\cite{lin2017focal}}{Anchor-based}
             & FPN
             & 36.5                      & 55.4 & 20.4
             & --                        & --   & --            \\
             & \aghlb{AGPF-S}
             & \aghlgain{37.9}{1.4}
             & \aghlb{57.2}
             & \aghlb{21.9}
             & --                        & --   & --            \\

            \methodpair{FCOS~\cite{tian2019fcos}}{Anchor-free}
             & FPN
             & 36.6                      & 56.0 & 21.0
             & --                        & --   & --            \\
             & \aghlb{AGPF-S}
             & \aghlgain{37.3}{0.7}
             & \aghlb{56.2}
             & \aghl{21.0}
             & --                        & --   & --            \\

            \methodpair{RepPoints~\cite{yang2019reppoints}}{Point-set}
             & FPN
             & 37.0                      & 56.7 & 20.4
             & --                        & --   & --            \\
             & \aghlb{AGPF-S}
             & \aghlgain{39.0}{2.0}
             & \aghlb{59.4}
             & \aghlb{23.1}
             & --                        & --   & --            \\

            \methodpair{ATSS~\cite{zhang2020bridging}}{Adaptive assignment}
             & FPN
             & 39.4                      & 57.6 & 23.6
             & --                        & --   & --            \\
             & \aghlb{AGPF-S}
             & \aghlgain{40.5}{1.1}
             & \aghlb{58.6}
             & \aghlb{24.1}
             & --                        & --   & --            \\

            \methodpair{GFL~\cite{li2020generalized}}{Quality-aware}
             & FPN
             & 40.2                      & 58.4 & \textbf{23.3}
             & --                        & --   & --            \\
             & \aghlb{AGPF-S}
             & \aghlgain{41.1}{0.9}
             & \aghlb{59.0}
             & \aghl{22.8}
             & --                        & --   & --            \\

            \methodpair{VFNet~\cite{zhang2021varifocalnet}}{IoU-aware}
             & FPN
             & 41.6                      & 59.5 & 24.4
             & --                        & --   & --            \\
             & \aghlb{AGPF-S}
             & \aghlgain{41.8}{0.2}
             & \aghl{59.5}
             & \aghlb{24.9}
             & --                        & --   & --            \\

            \methodpair{TOOD~\cite{feng2021tood}}{Task-aligned}
             & FPN
             & 42.4                      & 59.5 & 25.1
             & --                        & --   & --            \\
             & \aghlb{AGPF-S}
             & \aghlgain{43.2}{0.8}
             & \aghlb{60.7}
             & \aghlb{25.6}
             & --                        & --   & --            \\

            \midrule[0.55pt]
            \multicolumn{8}{@{}l}{
                \textit{Two-stage detection and instance segmentation (ResNet-50)}
            }                                                   \\

            \methodpair{Faster R-CNN~\cite{ren2015faster}}{Two-stage}
             & FPN
             & 37.4                      & 58.1 & 21.2
             & --                        & --   & --            \\
             & \aghlb{AGPF-B}
             & \aghlgain{39.8}{2.4}
             & \aghlb{61.0}
             & \aghlb{23.8}
             & --                        & --   & --            \\

            \methodpair{Dynamic R-CNN~\cite{zhang2020dynamic}}{Dynamic two-stage}
             & FPN
             & 38.9                      & 57.6 & 22.1
             & --                        & --   & --            \\
             & \aghlb{AGPF-B}
             & \aghlgain{40.9}{2.0}
             & \aghlb{60.0}
             & \aghlb{23.3}
             & --                        & --   & --            \\

            \methodpair{Grid R-CNN~\cite{lu2019grid}}{Grid-guided}
             & FPN
             & 40.4                      & 58.5 & 22.7
             & --                        & --   & --            \\
             & \aghlb{AGPF-B}
             & \aghlgain{42.2}{1.8}
             & \aghlb{61.0}
             & \aghlb{24.8}
             & --                        & --   & --            \\

            \methodpair{Cascade R-CNN~\cite{cai2018cascade}}{Cascade}
             & FPN
             & 40.3                      & 58.6 & 22.5
             & --                        & --   & --            \\
             & \aghlb{AGPF-B}
             & \aghlgain{42.9}{2.6}
             & \aghlb{61.9}
             & \aghlb{26.0}
             & --                        & --   & --            \\

            \methodpair{SABL R-CNN~\cite{wang2020side}}{Side-aware}
             & FPN
             & 41.0                      & 59.5 & 23.1
             & --                        & --   & --            \\
             & \aghlb{AGPF-B}
             & \aghlgain{44.0}{3.0}
             & \aghlb{61.9}
             & \aghlb{26.4}
             & --                        & --   & --            \\

            \multirow{4}{*}{%
            \makecell[l]{%
            Mask R-CNN~\cite{he2017mask}                        \\[-1.0pt]
            {\scriptsize\itshape Instance segmentation}%
            }%
            }
             & FPN
             & 38.2                      & 58.8 & 21.9
             & 34.7                      & 55.7 & 18.3          \\

             & FaPN~\cite{huang2021fapn}
             & 39.8                      & --   & 24.3
             & 36.4                      & --   & 18.1          \\

             & A$^2$-FPN~\cite{hu20212}
             & 40.0                      & --   & --
             & 36.2                      & 58.4 & {\bf 20.1}    \\

             & \aghlb{AGPF-B}
             & \aghlgain{41.1}{2.9}
             & \aghlb{62.0}
             & \aghlb{24.7}
             & \aghlgain{37.5}{2.8}
             & \aghlb{59.3}
             & \aghl{18.9}                                      \\

            \methodpair{Mask Scoring R-CNN~\cite{huang2019mask}}{Mask-quality scoring}
             & FPN
             & 38.2                      & 58.8 & 21.6
             & 36.0                      & 55.8 & 18.4          \\
             & \aghlb{AGPF-B}
             & \aghlgain{40.8}{2.6}
             & \aghlb{60.5}
             & \aghlb{24.2}
             & \aghlgain{38.6}{2.6}
             & \aghlb{58.7}
             & \aghlb{18.9}                                     \\

            \methodpair{Cascade Mask R-CNN~\cite{cai2018cascade}}{Cascade instance}
             & FPN
             & 41.2                      & 59.4 & 23.9
             & 36.0                      & 56.6 & 19.4          \\
             & \aghlb{AGPF-B}
             & \aghlgain{44.3}{3.1}
             & \aghlb{63.4}
             & \aghlb{26.9}
             & \aghlgain{39.4}{3.4}
             & \aghlb{60.8}
             & \aghlb{20.5}                                     \\

            \methodpair{HTC~\cite{chen2019hybrid}}{Hybrid cascade}
             & FPN
             & 42.3                      & 61.1 & 23.7
             & 37.4                      & 58.4 & 19.6          \\
             & \aghlb{AGPF-B}
             & \aghlgain{45.4}{3.1}
             & \aghlb{64.8}
             & \aghlb{27.5}
             & \aghlgain{40.8}{3.4}
             & \aghlb{62.3}
             & \aghlb{20.5}                                     \\

            \midrule[0.55pt]
            \multicolumn{8}{@{}l}{
                \textit{Backbone scaling with Cascade Mask R-CNN}
            }                                                   \\

            \multirow{2}{*}{ResNet-101}
             & FPN
             & 42.9                      & 61.0 & 24.4
             & 37.3                      & 58.2 & 19.7          \\
             & \aghlb{AGPF-B}
             & \aghlgain{44.6}{1.7}
             & \aghlb{63.3}
             & \aghlb{26.5}
             & \aghlgain{39.5}{2.2}
             & \aghlb{60.7}
             & \aghl{19.7}                                      \\

            \multirow{2}{*}{Swin-T}
             & FPN
             & 46.4                      & 65.8 & 29.0
             & 41.0                      & 63.2 & 21.4          \\
             & \aghlb{AGPF-B}
             & \aghlgain{47.3}{0.9}
             & \aghlb{67.1}
             & \aghlb{29.6}
             & \aghlgain{41.7}{0.7}
             & \aghlb{64.3}
             & \aghlb{21.8}                                     \\

            \multirow{2}{*}{Swin-B$^\dagger$}
             & FPN
             & 51.9                      & 70.5 & 35.4
             & 45.0                      & 68.1 & 28.9          \\
             & \aghlb{AGPF-B}
             & \aghlgain{54.5}{2.6}
             & \aghlb{73.3}
             & \aghlb{39.0}
             & \aghlgain{47.1}{2.1}
             & \aghlb{71.1}
             & \aghlb{29.0}                                     \\

            \bottomrule[1.35pt]
        \end{tabular}%
    }
    \caption{
        Neck replacement across detectors and backbones on COCO
        \texttt{val2017}.
        Small suffixes indicate AP gains over the corresponding
        FPN baselines; $^\dagger$ denotes $3\times$ multiscale training.
    }
    \label{tab:agpf_generalization}
    \vspace{-0.5em}
\end{table}

\section{Experiments}
\label{sec:experiments}
We evaluate AGPF on object detection, instance segmentation, and
panoptic segmentation on COCO, and on semantic segmentation on
Cityscapes and ADE20K.
Our COCO experiments first examine how individual output locations
select and use pyramid context, then compare AGPF with alternative
necks across detectors and backbones, and finally ablate its main
design choices.

\subsection{Experimental Setup}
\label{sec:experimental_setup}

\paragraph{Datasets and metrics}
We evaluate object detection and instance segmentation on
COCO val2017~\cite{lin2014microsoft} using box and mask AP.
Semantic segmentation is evaluated on
Cityscapes~\cite{cordts2016cityscapes} and
ADE20K~\cite{zhou2017scene} using mIoU and mAcc.
For panoptic segmentation on COCO, we report PQ,
PQ$^{\mathrm{Th}}$, PQ$^{\mathrm{St}}$, SQ, and
RQ~\cite{kirillov2019panoptic}.

\paragraph{Implementation}
Detection experiments use MMDetection~\cite{chen2019mmdetection}.
Controlled FPN--AGPF comparisons replace only the neck,
with the remaining model, training recipe, and evaluation
protocol unchanged.
Unless otherwise stated, analyses and ablations use
Cascade Mask R-CNN~\cite{cai2018cascade} with an
ImageNet-pretrained ResNet-50~\cite{he2016deep} and AGPF-S.
We use a feature width of $d=256$, following
FPN~\cite{lin2017feature}, and eight attention heads with an
FFN hidden width of $1024$, following
Deformable DETR~\cite{zhu2021iclr-deformable}.
AGPF-S, AGPF-B, and AGPF-L contain one, three, and six
fusion blocks, respectively.

The reference configuration uses a $12$-epoch ($1\times$)
schedule and a total batch size of $16$ on eight NVIDIA A6000 GPUs.
We use AdamW~\cite{loshchilov2017decoupled} with weight decay
$0.05$ and initial learning rates of $2\times10^{-4}$ for
ResNet and $1\times10^{-4}$ for Swin~\cite{liu2021swin} backbones.
For memory-design comparisons, detector latency and peak
training memory are measured with matched hardware, input size,
batch size.
Across repeated training runs without a fixed random seed,
the main evaluation metrics are reproducible within $0.1$ points.

\begin{figure}[t]
    \vspace{-.5em}
    \centering
    \includegraphics[width=0.95\linewidth]{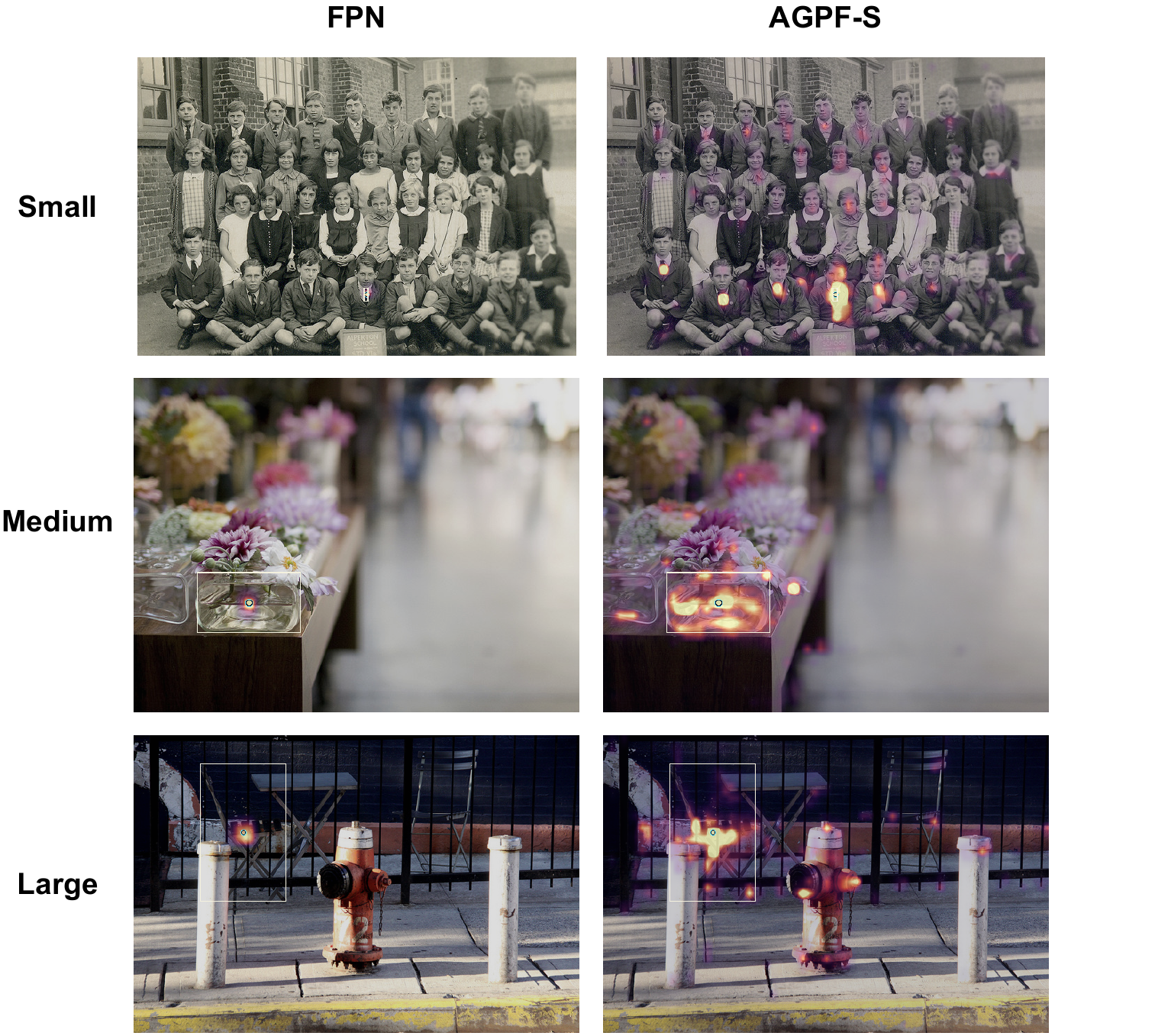}
    \caption{Cross-level dependency of FPN and AGPF-S across small, medium, and large COCO objects.}
    \label{fig:app_erf_scale_qualitative}
    \vspace{-1.5em}
\end{figure}

\begin{table}[t]
    \vspace{-0.5em}
    \centering
    \setlength{\tabcolsep}{2.5pt}
    \renewcommand{\arraystretch}{1.04}
    \small
    \resizebox{\columnwidth}{!}{%
        \begin{tabular}{llccc}
            \toprule[1.16pt]
            Method & Setting        & Recipe        & \apbox & \apmask             \\
            \midrule[0.60pt]

            \multicolumn{5}{l}{\textit{Conventional CNN-based systems}}            \\

            RetinaNet
                   & R50 + NASFPN
                   & 50e
                   & 40.5           & --                                           \\

            Mask R-CNN
                   & R101 + FPN
                   & MS, $3\times$
                   & 42.7           & 38.5                                         \\

            Cascade Mask R-CNN
                   & R101 + FPN
                   & MS, $3\times$
                   & 45.5           & 39.6                                         \\

            HTC
                   & R101 + FPN
                   & 20e
                   & 44.8           & 39.6                                         \\

            DetectoRS~\cite{qiao2021detectors}
                   & R101 + HTC
                   & 20e
                   & 50.5           & 43.9                                         \\

            \rowcolor{gray!15}
            \textbf{\ours{}-L}
                   & R50 + Cascade
                   & MS, $3\times$
                   & \textbf{51.1}  & \textbf{44.6}                                \\

            \midrule[0.47pt]
            \multicolumn{5}{l}{\textit{Modern detection and segmentation systems}} \\

            DETR~\cite{carion2020end}
                   & R50
                   & MS, 150e
                   & 39.9           & --                                           \\

            DAB-DETR~\cite{liu2022dabdetr}
                   & R50
                   & MS, 50e
                   & 42.3           & --                                           \\

            Deformable DETR~\cite{zhu2021iclr-deformable}
                   & R50, 2-stage
                   & MS, 50e
                   & 47.0           & --                                           \\

            Co-Deform-DETR~\cite{Zong_2023_ICCV}
                   & R50
                   & MS, $1\times$
                   & 49.5           & --                                           \\

            DINO~\cite{zhang2023iclr-dino}
                   & R50, 5-scale
                   & MS, $3\times$
                   & 51.2           & --                                           \\

            MaskFormer~\cite{cheng2021per}
                   & R50
                   & MS, 50e
                   & --             & 34.0                                         \\

            Mask2Former~\cite{cheng2022masked}
                   & R50
                   & LSJ, 50e
                   & 45.7           & 42.9                                         \\

            \rowcolor{gray!15}
            \textbf{\ours{}-L + HTC}
                   & R50
                   & MS, $3\times$
                   & \textbf{52.4}  & \textbf{45.8}                                \\

            Mask DINO~\cite{li2023mask}
                   & R50
                   & LSJ, 50e
                   & 50.5           & 46.0                                         \\

            Co-DETR~\cite{Zong_2023_ICCV}
                   & R50, DINO++
                   & LSJ, $3\times$
                   & 54.8           & --                                           \\

            \rowcolor{gray!15}
            \textbf{\ours{}-L + HTC++}
                   & R50
                   & LSJ, 50e
                   & \textbf{55.2}  & \textbf{48.0}                                \\

            \bottomrule[1.16pt]
        \end{tabular}%
    }
    \caption{
        Comparison with representative detection and instance-segmentation
        methods on COCO. MS indicates multi-scale training, and LSJ denotes large-scale jittering~\cite{ghiasi2021simple}.
    }
    \label{tab:modern_method_comparison}
    \vspace{-0.5em}
\end{table}

\begin{table}[b]
    \vspace{-0.5em}
    \centering
    \setlength{\tabcolsep}{2.45pt}
    \renewcommand{\arraystretch}{1.05}
    \resizebox{\linewidth}{!}{%
        \begin{tabular}{llcccc}
            \toprule[1.22pt]
            \textbf{Method}
             & \textbf{\boldmath$\#P_{\text{Enc.}}$}
             & \textbf{\boldmath Hours}
             & \textbf{\boldmath Iter. (s)}
             & \textbf{\boldmath Train Mem.}
             & \textbf{\boldmath$\mathrm{AP}_{B/M}$} \\
            \midrule[0.68pt]

            Mask2Former
             & 29.5M
             & 62
             & 0.6
             & 10732
             & 45.7 / 42.9                           \\

            Mask DINO
             & 37.6M
             & 102
             & 1.0
             & 14520
             & 50.5 / 46.0                           \\

            DINO
             & 37.1M
             & 83
             & 1.1
             & 24394
             & 51.2 / \;\;--\;\;                     \\

            Co-DETR
             & 41.0M
             & 132
             & 1.8
             & 29228
             & 54.8 / \;\;--\;\;                     \\

            \textbf{AGPF-HTC}
             & 34.0M
             & 83
             & 1.1
             & 16972
             & 52.4 / 45.8                           \\

            \textbf{AGPF-HTC++}
             & 34.0M
             & 169
             & 1.6
             & 15707
             & \textbf{55.2 / 48.0}                  \\

            \bottomrule[1.22pt]
        \end{tabular}%
    }
    \caption{
        System-level accuracy and efficiency comparison.
        $\#P_{\rm Enc.}$ counts backbone and encoder/fusion parameters;
        Hours, Iter., and Mem. denote training hours, seconds per iteration,
        and peak training memory (MB), respectively.
    }
    \label{tab:system_efficiency}
    \vspace{-0.5em}
\end{table}

\begin{table*}[t]
    \vspace{-0.5em}
    \centering
    \small

    \makebox[\textwidth][c]{%
        \begin{minipage}{1.04\textwidth}
            \centering

            \begin{tabular}{@
                        {}c
                    @{\hspace{0.015\linewidth}}
                    c
                    @{\hspace{0.015\linewidth}}
                    c@{}
                }

                \begin{minipage}[t]{0.350\linewidth}
                    \vspace{0pt}
                    \centering
                    \renewcommand{\arraystretch}{1.10}
                    \setlength{\tabcolsep}{2.45pt}

                    \resizebox{\linewidth}{!}{%
                        \begin{tabular}{@{}lccccccc@{}}
                            Method
                             & P2 Pool
                             & \#Layers
                             & \apbox
                             & \apmask
                             & Params
                             & Mem.
                             & Iter. (s)                       \\
                            \midrule[1.10457pt]

                            \ours-T
                             & \cmark    & 1     & 42.1 & 37.6
                             & 6.49M     & 5520  & 0.41        \\

                            \ours-S
                             & \xmark    & 1     & 42.9 & 38.3
                             & 6.49M     & 6389  & 0.51        \\

                            \ours-B
                             & \xmark    & 3     & 44.3 & 39.4
                             & 8.07M     & 9392  & 0.62        \\

                            \ours-L
                             & \xmark    & 6     & 44.6 & 39.5
                             & 10.45M    & 13963 & 0.94        \\

                            \ours-L$^{\dagger}$
                             & \xmark    & 6     & 50.9 & 44.3
                             & 10.45M    & 13963 & 0.94        \\
                        \end{tabular}%
                    }
                \end{minipage}
                 &

                \begin{minipage}[t]{0.335\linewidth}
                    \vspace{0pt}
                    \centering
                    \renewcommand{\arraystretch}{1.06}
                    \setlength{\tabcolsep}{2.55pt}

                    \resizebox{\linewidth}{!}{%
                        \begin{tabular}{@{}lcccc@{}}
                            Position design
                             & Level Emb.
                             & Feature-aligned
                             & \apbox
                             & \apmask                  \\
                            \midrule[1.02904pt]

                            No position
                             & \xmark          & \xmark
                             & 41.3            & 37.4   \\

                            Fixed 2D sin-cos
                             & \cmark          & \xmark
                             & 41.8            & 37.7   \\

                            Local RoPE
                             & \cmark          & \xmark
                             & 41.5            & 37.4   \\

                            \farrope{}
                             & \cmark          & \cmark
                             & 42.7            & 38.2   \\

                            Mixed \farrope{}
                             & \cmark          & \cmark
                             & \textbf{42.9}
                             & \textbf{38.3}            \\
                        \end{tabular}%
                    }
                \end{minipage}
                 &

                \begin{minipage}[t]{0.275\linewidth}
                    \vspace{0pt}
                    \centering
                    \renewcommand{\arraystretch}{1.17}
                    \setlength{\tabcolsep}{2.45pt}

                    \vspace{-0.1pt}

                    \resizebox{\linewidth}{!}{%
                        \begin{tabular}{@{}lccccc@{}}
                            Method
                             & Global
                             & KV Pool
                             & Fused
                             & \apbox
                             & \apmask                                \\
                            \midrule[1.15632pt]

                            FPN
                             & \xmark        & --            & --
                             & 41.2          & 36.0                   \\

                            Full attention
                             & \cmark        & \xmark        & \cmark
                             & 42.7          & 38.2                   \\

                            Enforce Pool
                             & \cmark        & \cmark        & \cmark
                             & 42.4          & 37.9                   \\

                            \ours{} w/ mask
                             & \cmark        & \cmark        & \xmark
                             & 42.8          & 38.2                   \\

                            \ours{}
                             & \cmark        & \cmark        & \cmark
                             & \textbf{42.9} & \textbf{38.3}          \\
                        \end{tabular}%
                    }
                \end{minipage}
                \\[-1pt]

                {\scriptsize (a) Model scaling}
                 &
                {\scriptsize (b) Positional encoding}
                 &
                {\scriptsize (c) Attention components}
            \end{tabular}

        \end{minipage}%
    }

    \caption{
        Controlled ablation studies of the main AGPF design choices.
        $^{\dagger}$ denotes $3\times$ multi-scale training.
    }
    \label{tab:agpf_ablations}
    \vspace{-1em}
\end{table*}

\subsection{Experiments on COCO}
\label{sec:coco}

\paragraph{Pyramid source selection}
We analyze AGPF-S over all $5{,}000$ COCO \texttt{val2017}
images, sampling a query at the bounding-box center of each of
$36{,}334$ valid objects at every output level.
Routing statistics follow Eq.~\ref{eq:routing_mass}, with
enrichment computed per image before aggregation.

With the default pooled memory, $P_2$ queries assign approximately
$75\%$ of their attention to $P_4$, two levels away, and $92\%$
to $P_4$ and $P_5$ together (Fig.~\ref{fig:analysis}(b)).
This preference persists after accounting for memory-token counts:
the $P_2$-to-$P_4$ allocation is $2.48\times$ the uniform-token
reference (Supp. C.2).
The unpooled reference also favors coarse sources
(Fig.~\ref{fig:analysis}(b)), indicating that this preference is
not specific to memory compression.
A fine output resolution therefore does not imply a preference
for fine or adjacent source levels.

\paragraph{Routing across object scales}
At each query level, AGPF-S queries associated with larger
objects allocate more attention to $P_5$ and less to $P_4$
(Fig.~\ref{fig:analysis}(b)).
For $P_2$ queries, attention to $P_5$ increases from $16.0\%$
for small objects to $18.9\%$ for large ones, while attention
to $P_4$ decreases from $75.8\%$ to $72.8\%$.
Across the five query levels, the large-minus-small increase
in $P_5$ attention ranges from $2.8$ to $4.6$ percentage points,
with all image-bootstrap $95\%$ confidence intervals above zero
(Supp.~C.4)
Finer object-scale bins show the same overall trend
(Supp.~C.3).
These differences occur within the same output grid:
a query's output resolution alone does not describe which
pyramid sources it favors.

\begin{figure}[h]
    \vspace{-0.3em}
    \centering
    \includegraphics[width=0.85\linewidth]{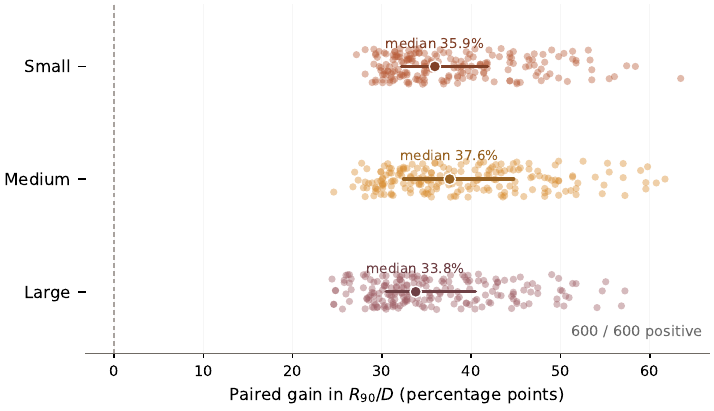}
    \caption{Paired change in normalized $R_{90}$ from FPN to AGPF-S.}
    \label{fig:paired_r90_delta}
    \vspace{-1em}
\end{figure}

\paragraph{Spatial dependency}
Source-level routing describes how a query distributes attention
across pyramid levels; gradient sensitivity identifies which
input-pyramid locations affect its output.
We compare FPN and AGPF-S on the same $600$ object-centered
queries, with $200$ objects from each COCO size group.
The median $R_{90}/D$ increases from $2.3\%$ for FPN to
$38.3\%$ for AGPF-S, with a larger radius for AGPF-S in all $600$ pairs
(Fig.~\ref{fig:paired_r90_delta}).
AGPF-S also retains a median of $47.1\%$ of its gradient mass
within $0.05D$ of the query, whereas FPN accumulates nearly
all its mass within this neighborhood
(Fig.~\ref{fig:positional_consistency}, left;
Fig.~\ref{fig:app_erf_scale_qualitative}).
These dependencies complement the routing results:
AGPF selects context across levels and exhibits sensitivity
to spatially distant pyramid regions while retaining
a substantial local component.

\paragraph{Translation consistency}
We compare level-local coordinates and FA-RoPE on $200$ COCO
images using one checkpoint.
Both coordinate schemes use the same extracted queries and keys
for each input.
For query--memory pairs with different effective strides,
FA-RoPE reduces mean attention drift by $93.6$--$94.9\%$
across translations of $64$, $128$, and $192$ pixels
(Fig.~\ref{fig:positional_consistency}, right).
A complementary control that translates the extracted features
directly preserves attention to numerical precision with FA-RoPE
(Fig.~\ref{fig:panel_c_supp_controlled_feature_shift}).
In the positional ablation, replacing level-local RoPE with
fixed-frequency FA-RoPE improves box AP by $1.2$ and mask AP
by $0.8$, with learned level embeddings retained
(Table~\ref{tab:agpf_ablations}(b)).
These results support a shared spatial reference for joint
retrieval across pyramid resolutions.

\paragraph{Comparison with pyramid-fusion necks}
Table~\ref{tab:neck_detector_comparison} compares AGPF-S with
alternative necks under matched Faster R-CNN, Mask R-CNN, and
Cascade Mask R-CNN settings.
With a single joint-retrieval block, AGPF-S achieves the highest
overall box AP in all three detector groups and the highest
mask AP in both instance-segmentation groups.
Compared with AugFPN, the strongest alternative in overall AP,
AGPF-S improves box AP by $0.1$--$0.8$ points and mask AP by
$0.8$--$1.8$ points.
Joint source selection thus offers an effective alternative
to fusion through prescribed pyramid pathways.
Small-object mask AP remains slightly below FPN, possibly
reflecting limited refinement of fine spatial detail within
a single fusion block.

\paragraph{Across detectors and backbones}
We use the single-block AGPF-S with one-stage heads and the
deeper AGPF-B with the more complex two-stage and cascade
prediction pipelines (Table~\ref{tab:agpf_generalization}).
Replacing FPN improves overall box AP in all $19$ evaluated
configurations and mask AP in all seven instance-segmentation
configurations.
AGPF-B also surpasses FPN in small-object mask AP with both
Mask R-CNN and Cascade Mask R-CNN.

The improvements span different ways of consuming pyramid features.
AGPF-S increases RepPoints box AP from $37.0$ to $39.0$,
while AGPF-B improves HTC by $3.1$ box AP and $3.4$ mask AP.
The gains persist with ResNet-101, Swin-T, and Swin-B backbones.
With Swin-B under matched $3\times$ multiscale training,
Cascade Mask R-CNN improves from $51.9/45.0$ to $54.5/47.1$
box/mask AP.
Across these architectures, the same retrieval operation
improves prediction while preserving the pyramid interface
expected by their existing heads.

\begin{table}[t]
    \centering
    \fontsize{7.5pt}{8.0pt}\selectfont
    \setlength{\tabcolsep}{6.5pt}
    \begin{tabular}{@{}lccccc@{}}
        Ablation                   & Setting     & Default & \apbox        & \apmask       & Mem. \\
        \midrule[0.70pt]
        \multirow{4}{*}{\#Heads}   & 2           &         & 42.6          & 38.1          & 6383 \\
                                   & 4           &         & 42.5          & 38.0          & 6356 \\
                                   & 8           & \cmark  & \textbf{42.9} & \textbf{38.3} & 6389 \\
                                   & 16          &         & \textbf{42.9} & \textbf{38.3} & 6405 \\
        \midrule
        \multirow{3}{*}{Pooling}   & AvgPool     &         & 42.7          & 38.1          & 6392 \\
                                   & MaxPool     &         & 42.8          & \textbf{38.3} & 6387 \\
                                   & DWConv      & \cmark  & \textbf{42.9} & \textbf{38.3} & 6389 \\
        \midrule
        \multirow{4}{*}{KV stride} & [8,4,2,1,1] &         & 41.5          & 37.3          & 6172 \\
                                   & [4,2,1,1,1] & \cmark  & \textbf{42.9} & \textbf{38.3} & 6389 \\
                                   & [4,2,2,1,1] &         & 41.7          & 37.5          & 6356 \\
                                   & [4,4,2,2,1] &         & 41.7          & 37.5          & 6331 \\
    \end{tabular}
    \caption{Single-layer AGPF design ablations.}
    \label{tab:single_pass_design}
    \vspace{-1em}
\end{table}

\paragraph{Comparison with complete systems}
Table~\ref{tab:modern_method_comparison} compares complete
AGPF-based detectors with representative detection and
instance-segmentation systems.
Here, AGPF-L is combined with enhanced prediction heads
and training recipes; configuration details are provided
in Supp.~F.

With ResNet-50, AGPF-L reaches $51.1/44.6$ box/mask AP
with Cascade Mask R-CNN and $52.4/45.8$ with HTC.
The stronger HTC++ configuration, trained for $50$ epochs
with large-scale jittering, reaches $55.2/48.0$,
exceeding the listed Co-DETR result by $0.4$ box AP
and Mask DINO by $2.0$ mask AP.

Table~\ref{tab:system_efficiency} reports the corresponding training
costs under matched measurement conditions.
Compared with Co-DETR, AGPF-HTC++ uses $46.3\%$ less peak training
memory and takes $1.6$ rather than $1.8$ seconds per iteration,
although its longer schedule requires $169$ hours in total versus
$132$; the higher accuracy therefore does not imply a uniformly lower
training cost.
AGPF can thus serve as the fusion component of competitive
region-based detection and segmentation systems, alongside stronger
heads and training recipes.

\paragraph{Panoptic segmentation}
We replace FPN with AGPF in Panoptic FPN~\cite{kirillov2019panoptic}
and K-Net~\cite{zhang2021k}, retaining their prediction components
(Table~\ref{tab:panoptic_segmentation}).
With ResNet-50, AGPF-B improves Panoptic FPN from $42.5$ to
$46.4$ PQ and K-Net from $47.1$ to $49.0$ PQ.
With Swin-L, AGPF-L improves K-Net from $54.5$ to $55.8$ PQ.
Both thing and stuff PQ increase in all three settings.
For Panoptic FPN, the gains are $3.1$ PQ$^{\mathrm{Th}}$
and $5.1$ PQ$^{\mathrm{St}}$, showing that the retrieved
pyramid context benefits both individual objects and semantic
regions within the same prediction system.

\begin{table}[t]
    \vspace{-0.5em}
    \centering
    \setlength{\tabcolsep}{5.0pt}
    \renewcommand{\arraystretch}{1.08}
    \small
    \resizebox{\linewidth}{!}{%
        \begin{tabular}{lllcccc}
            \toprule[1.16pt]
            \multirow{2}{*}{Framework}
             & \multirow{2}{*}{Backbone}
             & \multirow{2}{*}{Fusion}
             & \multicolumn{2}{c}{Cityscapes}
             & \multicolumn{2}{c}{ADE20K}                                                 \\
            \cmidrule[0.40pt](lr){4-5}
            \cmidrule[0.40pt](lr){6-7}
             &                                &               & mIoU & mAcc & mIoU & mAcc \\
            \midrule[0.60pt]

            \multirow{4}{*}{UPerNet~\cite{xiao2018unified}}
             & \multirow{2}{*}{R50}
             & FPN
             & 79.3                           & 87.2          & 40.7 & 51.9               \\
             &                                & {\bf AGPF-B}
             & \textbf{80.0}                  & \textbf{87.4}
             & \textbf{43.7}                  & \textbf{54.7}                             \\

            \cmidrule[0.51pt](lr){2-7}

             & \multirow{2}{*}{Swin-L}
             & FPN
             & 82.3                           & 88.8          & 52.0 & 63.2               \\
             &                                & {\bf AGPF-L}
             & \textbf{82.5}                  & \textbf{89.0}
             & \textbf{52.4}                  & \textbf{63.7}                             \\

            \midrule[0.51pt]

            \multirow{4}{*}{K-Net~\cite{zhang2021k}}
             & \multirow{2}{*}{R50}
             & FPN
             & 80.5                           & 87.1          & 43.4 & 54.8               \\
             &                                & {\bf AGPF-B}
             & \textbf{80.6}                  & \textbf{87.4}
             & \textbf{44.7}                  & \textbf{55.6}                             \\

            \cmidrule[0.51pt](lr){2-7}

             & \multirow{2}{*}{Swin-L}
             & FPN
             & 82.3                           & 88.5          & 52.2 & 63.5               \\
             &                                & {\bf AGPF-L}
             & \textbf{83.5}                  & \textbf{89.5}
             & \textbf{52.5}                  & \textbf{63.8}                             \\

            \bottomrule[1.16pt]
        \end{tabular}%
    }
    \caption{Semantic segmentation on Cityscapes and ADE20K.}
    \label{tab:semantic_segmentation}
    \vspace{-1em}
\end{table}

\subsection{Semantic Segmentation}
\label{sec:semantic_segmentation}

We evaluate AGPF in UPerNet~\cite{xiao2018unified} and
K-Net~\cite{zhang2021k} on Cityscapes and ADE20K
(Table~\ref{tab:semantic_segmentation}).
We replace their pyramid-fusion modules with AGPF-B for
ResNet-50 and AGPF-L for Swin-L, keeping the remaining
model and training settings unchanged.
Both mIoU and mAcc improve in every paired comparison.
With UPerNet and ResNet-50, mIoU increases from $79.3$
to $80.0$ on Cityscapes and from $40.7$ to $43.7$ on ADE20K.
The gains also extend to Transformer backbones: K-Net with
Swin-L improves from $82.3$ to $83.5$ mIoU on Cityscapes.
Joint pyramid retrieval thus supplies useful context for
pixelwise scene labeling as well as object-level prediction,
without requiring a task-specific fusion structure.

\subsection{Ablation Studies}
\label{sec:ablations}

\paragraph{Memory resolution}
The default pooled memory preserves the accuracy of the
full-memory reference: $42.9/38.3$ versus $42.7/38.2$
box/mask AP (Table~\ref{tab:agpf_ablations}(c)).
Further compression, however, affects accuracy
(Table~\ref{tab:single_pass_design}).
Changing the KV strides from $[4,2,1,1,1]$ to
$[4,2,2,1,1]$, which additionally downsamples $P_4$,
reduces performance to $41.7/37.5$.
The more aggressive $[8,4,2,1,1]$ configuration obtains
$41.5/37.3$.
The sensitivity to $P_4$ compression is consistent with its
prominent role in retrieval and supports retaining its
native memory resolution while compressing the finer levels.

\paragraph{Fusion depth and query density}
Table~\ref{tab:agpf_ablations}(a) compares the AGPF variants.
Increasing fusion depth from one block to three improves
box/mask AP by $1.4/1.1$.
Increasing it further to six blocks adds only $0.3/0.1$ AP,
while training iteration time rises from $0.62$ to $0.94$
seconds.
AGPF-B therefore captures most of the depth-related gain
at a lower training cost than AGPF-L.

AGPF-T instead reduces query density by pooling $P_2$ queries.
Relative to AGPF-S, it saves $13.6\%$ of training memory
and reduces iteration time from $0.51$ to $0.41$ seconds,
but loses $0.8/0.7$ box/mask AP.
We retain dense queries in the standard S/B/L variants,
with AGPF-T providing an additional option when memory
is constrained.

\paragraph{Positional encoding}
Both feature-aligned variants outperform the unaligned
encodings in Table~\ref{tab:agpf_ablations}(b).
Within the shared image coordinate system, mixed RoPE
adds $0.2$ box AP and $0.1$ mask AP over fixed-frequency
FA-RoPE and is used by default.
The larger improvement is already obtained with the shared
spatial reference, consistent with the translation analysis
in Sec.~\ref{sec:coco}.

\begin{table}[t]
    \vspace{-0.5em}
    \centering
    \setlength{\tabcolsep}{2.8pt}
    \renewcommand{\arraystretch}{1.08}
    \small
    \resizebox{\linewidth}{!}{%
        \begin{tabular}{@{}lllccccc@{}}
            \toprule[1.14pt]
            Framework
             & Backbone
             & Fusion
             & $\mathrm{PQ}$
             & $\mathrm{PQ}^{Th}$
             & $\mathrm{PQ}^{St}$
             & $\mathrm{SQ}$
             & $\mathrm{RQ}$        \\
            \midrule[0.59pt]

            \multirow{2}{*}{Panoptic FPN~\cite{kirillov2019panoptic}}
             & \multirow{2}{*}{R50}
             & FPN
             & 42.5
             & 50.3
             & 30.7
             & 78.1
             & 51.7                 \\
             &
             & \textbf{AGPF-B}
             & \textbf{46.4}
             & \textbf{53.4}
             & \textbf{35.8}
             & \textbf{81.4}
             & \textbf{55.8}        \\

            \midrule[0.50pt]

            \multirow{4}{*}{K-Net~\cite{zhang2021k}}
             & \multirow{2}{*}{R50}
             & FPN
             & 47.1
             & 51.6
             & 40.2
             & 80.6
             & 57.3                 \\
             &
             & \textbf{AGPF-B}
             & \textbf{49.0}
             & \textbf{53.4}
             & \textbf{42.2}
             & \textbf{81.4}
             & \textbf{59.0}        \\

            \cmidrule[0.50pt](l){2-8}

             &
            \multirow{2}{*}{Swin-L}
             & FPN
             & 54.5
             & 60.2
             & 46.0
             & 82.3
             & 65.4                 \\
             &
             & \textbf{AGPF-L}
             & \textbf{55.8}
             & \textbf{61.6}
             & \textbf{46.9}
             & \textbf{82.6}
             & \textbf{66.7}        \\

            \bottomrule[1.14pt]
        \end{tabular}%
    }
    \caption{
        Panoptic segmentation on COCO \texttt{val2017}.
    }
    \label{tab:panoptic_segmentation}
    \vspace{-1em}
\end{table}

\paragraph{Pooling operator}
Average and max pooling remain within $0.2$ AP of
depthwise-convolution pooling
(Table~\ref{tab:single_pass_design}).
The gains therefore do not require a learnable pooling
operator.
We use depthwise convolution, which gives the highest
box AP and ties max pooling in mask AP.
The Enforce Pool variant obtains $42.4/37.9$
(Table~\ref{tab:agpf_ablations}(c)), below the default
$42.9/38.3$.
We therefore skip pooling at stride one, where no spatial
reduction is required.

\paragraph{Attention configuration}
Varying the number of heads from two to sixteen yields
$42.5$--$42.9$ box AP and $38.0$--$38.3$ mask AP
(Table~\ref{tab:single_pass_design}).
Eight and sixteen heads both obtain $42.9/38.3$;
we use eight.
Explicit padding masking with the non-fused implementation
obtains $42.8/38.2$
(Table~\ref{tab:agpf_ablations}(c)).
We retain the fused implementation, which achieves
comparable accuracy without an explicit padding mask.

\section{Discussion}
\label{sec:discussion}

\noindent\textbf{Retrieval preference and spatial detail.}
Low average attention to fine-resolution memory does not imply that
fine-resolution features are dispensable: a source used sparingly in
aggregate may still matter to particular predictions, and
fine-resolution features define dense output queries as well as
supplying context.
AGPF compresses the memory while preserving these queries, and the
accuracy loss from additionally pooling queries
(Sec.~\ref{sec:ablations}) separates the two roles.
Reducing context density and reducing the spatial resolution of
prediction features are therefore distinct tradeoffs.

\noindent\textbf{Scope and future work.}
AGPF addresses pyramid fusion within the interface used by existing
dense prediction heads.
In DETR-style systems, encoder representations are designed together
with object-query decoding and the set-prediction
objective~\cite{carion2020end,zhu2021iclr-deformable}, so their
effectiveness within that joint design does not establish their
effectiveness as standalone pyramid necks serving other prediction
heads; we therefore do not treat them as interchangeable neck
baselines, and Tables~\ref{tab:modern_method_comparison}
and~\ref{tab:system_efficiency} compare complete-system accuracy and
cost rather than encoder quality in isolation.
Extending compact pyramid retrieval to such joint encoder--decoder
designs remains future work.

\section{Conclusion}
\label{sec:conclusion}

This work shows that an output location's level in a feature pyramid
does not fully describe its use of context: when pyramid sources are
jointly accessible, queries can favor nonadjacent levels, and their
allocation across sources changes with object scale even at a fixed
query level.
AGPF turns this observation into a practical fusion design in which
dense output queries retrieve from a compact multilevel memory with a
shared spatial reference, and controlled neck replacements improve
detection and segmentation across the evaluated architectures while
retaining their prediction interfaces.
Dense prediction outputs can therefore be supported by selective
retrieval from compact context, which suggests treating the output
hierarchy and the organization of context as separate design choices,
and using the source preferences of individual locations to guide how
pyramid information is made available and combined.

    {
        \small
        \bibliographystyle{ieeenat_fullname}
        \bibliography{main}
    }

\clearpage
\appendix
\raggedbottom

\begin{figure*}[!b]
    \centering
    \includegraphics[width=\textwidth]
    {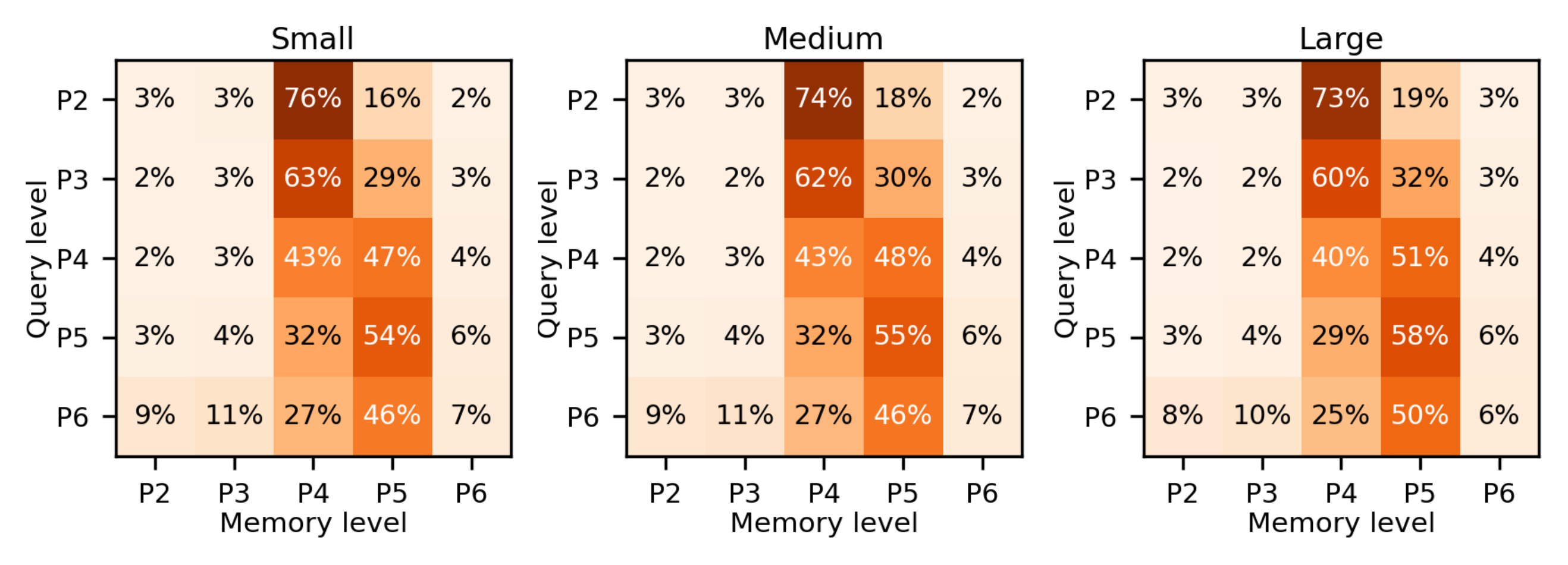}
    \caption{
        Source allocation for small, medium, and large
        COCO instances.
        Each entry is the mean attention mass from a fixed
        query level to a memory level, averaged over instances
        in the corresponding size group and attention heads.
    }
    \label{fig:Scale-conditioned-routing}
\end{figure*}

\section{Measurement Protocols}
\label{app:analysis_protocols}

We use the measurements defined in Sec.~\ref{sec:approach}.
All analyses are conducted on COCO \texttt{val2017}.
Unless otherwise stated, routing and translation analyses use
AGPF-S with its default pooled memory.

\subsection{Object-centered query sampling}

We evaluate routing on all $5{,}000$ COCO
images and $36{,}334$ valid ground-truth instances.
For each instance, we sample a query at its bounding-box center
at every output level, allowing object scales to be compared
while holding the query level fixed.
Attention mass is aggregated over instances and attention heads
following Eq.~\ref{eq:routing_mass}.
Enrichment uses the memory-token counts of each image before
aggregation.
Small, medium, and large instances follow the COCO object-size
groups.

\subsection{Matched gradient measurements}

We sample $600$ objects, with $200$ from each COCO size group
and at most one object per image.
Each query is placed at the ground-truth box center and assigned
to a pyramid level according to object scale after test-time
resizing.
FPN and AGPF-S are evaluated at the same query location and
level for each object.
We exclude objects whose centers lie within $5\%$ of an image
boundary to reduce truncation of the dependency maps.
All radius statistics use the normalized gradient distribution
defined in Sec.~\ref{sec:approach}.

\subsection{Matched input translations}

We evaluate $200$ COCO images under translations of
$64$, $128$, and $192$ pixels.
These displacements are integer-aligned with all effective
query and memory strides.
Features are recomputed after each input translation, and
attention is compared at corresponding interior query locations.
For each input, level-local coordinates and FA-RoPE use the
same checkpoint and extracted query/key features;
only their positional coordinates differ.
Drift is measured using the source-normalized attention
distributions and normalized Jensen--Shannon divergence
defined in Sec.~\ref{sec:approach}.

\section{Parameter-matched FPN}
\label{app:neck_capacity}

\begin{table}[H]
    \centering
    \small
    \setlength{\tabcolsep}{4pt}
    \begin{tabular}{@{}lccc@{}}
        \toprule
        Neck & Params (M) & Box AP        & Mask AP       \\
        \midrule
        FPN
             & 3.34       & 41.2          & 36.0          \\
        FPN, expanded
             & 6.49       & 41.2          & 36.1          \\
        AGPF-S
             & 6.49       & \textbf{42.9} & \textbf{38.3} \\
        \bottomrule
    \end{tabular}
    \caption{
        Neck-capacity comparison with Cascade Mask R-CNN
        and ResNet-50 under the reference training setting.
        Parameters count the neck only.
    }
    \label{tab:source_selection_controls}
\end{table}

We expand FPN to approximately match the neck parameter
count of AGPF-S.
The expanded FPN achieves $41.2/36.1$ box/mask AP,
compared with $42.9/38.3$ for AGPF-S
(Table~\ref{tab:source_selection_controls}).
At comparable capacity, AGPF-S retains gains of
$1.7$ box AP and $2.2$ mask AP over the expanded FPN.

\section{Checks on Pyramid Retrieval}
\label{app:source_selection}

\subsection{Object scale at fixed query resolution}
\label{app:fixed_query_resolution}

Figure~\ref{fig:Scale-conditioned-routing} gives the complete
source-allocation matrices for the three COCO size groups.
Fine-level queries favor $P_4$, whereas coarser query levels
place more attention on $P_5$.
Within every fixed query level, larger objects shift attention
from $P_4$ toward $P_5$.
The scale-dependent shift therefore does not result from
assigning larger objects to coarser output levels:
source allocation changes even when output resolution is held fixed.

\subsection{Accounting for token availability}

\begin{figure}[H]
    \centering
    \includegraphics[width=.68\linewidth]
    {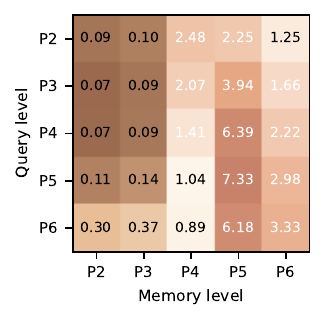}
    \caption{
        Source-level enrichment relative to uniform attention
        over memory tokens.
        Rows denote query levels and columns denote memory
        levels.
        A value of one matches the uniform-token reference.
    }
    \label{fig:routing_enrichment}
\end{figure}

For $P_2$ queries, enrichment is $2.48$ for $P_4$ memory
but only $0.09$ for $P_2$ memory
(Fig.~\ref{fig:routing_enrichment}).
The coarse-source preference persists after accounting
for the number of tokens supplied by each level.

\subsection{Finer scale partitions}

\begin{figure}[H]
    \centering
    \includegraphics[width=.82\linewidth]
    {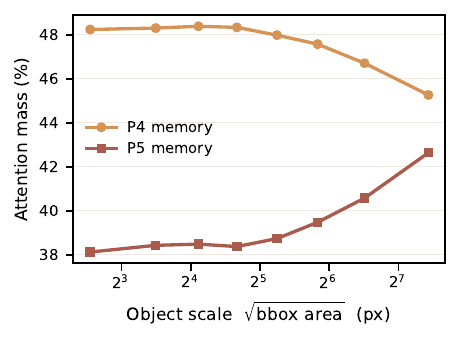}
    \caption{
        Source allocation across eight equal-frequency
        object-scale bins.
        Scale is the square root of bounding-box area;
        attention is averaged over instances within each bin,
        with equal weighting of query levels and attention heads.
    }
    \label{fig:routing_continuous_scale}
\end{figure}

From the smallest to the largest bin, mean attention to $P_4$
decreases from $48.2\%$ to $45.3\%$, while attention to $P_5$
increases from $38.1\%$ to $42.6\%$.
The smallest bins vary little; the shift toward $P_5$ becomes
clearer for larger objects.
The trend thus extends beyond the three COCO size groups.

\subsection{Uncertainty under image resampling}

\begin{figure}[H]
    \centering
    \includegraphics[width=.94\linewidth]
    {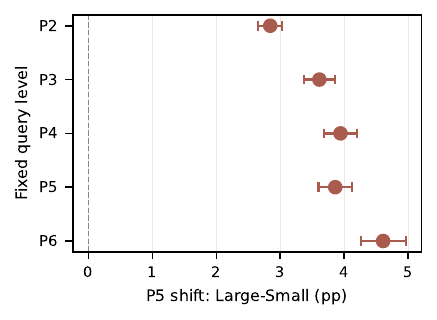}
    \caption{
        Large-minus-small change in attention to $P_5$ at
        each fixed query level.
        Points show the change in percentage points, and
        error bars show $95\%$ image-level bootstrap
        confidence intervals.
        The dashed line marks zero change.
    }
    \label{fig:routing_bootstrap_ci}
\end{figure}

We estimate uncertainty in the large-minus-small change in
$P_5$ attention by resampling complete images, keeping instances
from the same image together.
This accounts for correlations among objects within an image.
At every fixed query level, the $95\%$ confidence interval lies
above zero (Fig.~\ref{fig:routing_bootstrap_ci}).
For the evaluated checkpoint, the increase in $P_5$ allocation
remains positive under image-level resampling.

\subsection{Differences among attention heads}

\begin{figure}[H]
    \centering
    \includegraphics[width=.90\linewidth]
    {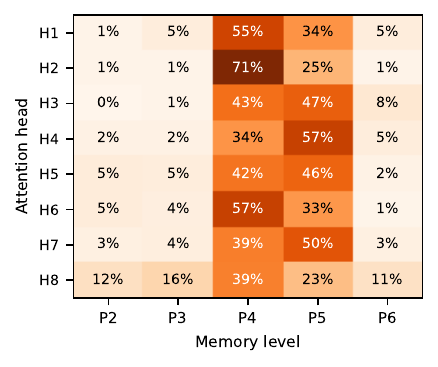}
    \caption{
        Source allocation of individual attention heads in
        AGPF-S.
        Each row reports attention mass averaged over objects
        and query levels.
    }
    \label{fig:routing_head_specialization}
\end{figure}

The aggregate routing pattern combines heads with different
source preferences (Fig.~\ref{fig:routing_head_specialization}).
For example, head H2 assigns $71\%$ of its attention to $P_4$,
whereas head H4 assigns $57\%$ to $P_5$.
Head H8 distributes attention more broadly and allocates more
mass to the finer levels than the other heads.
The average allocation therefore summarizes several distinct
retrieval patterns within the same block.

\section{Robustness of Spatial Dependence}
\label{app:spatial_dependency}

\subsection{Grouping and radius thresholds}

\begin{figure}[!htb]
    \centering
    \includegraphics[width=.94\linewidth]
    {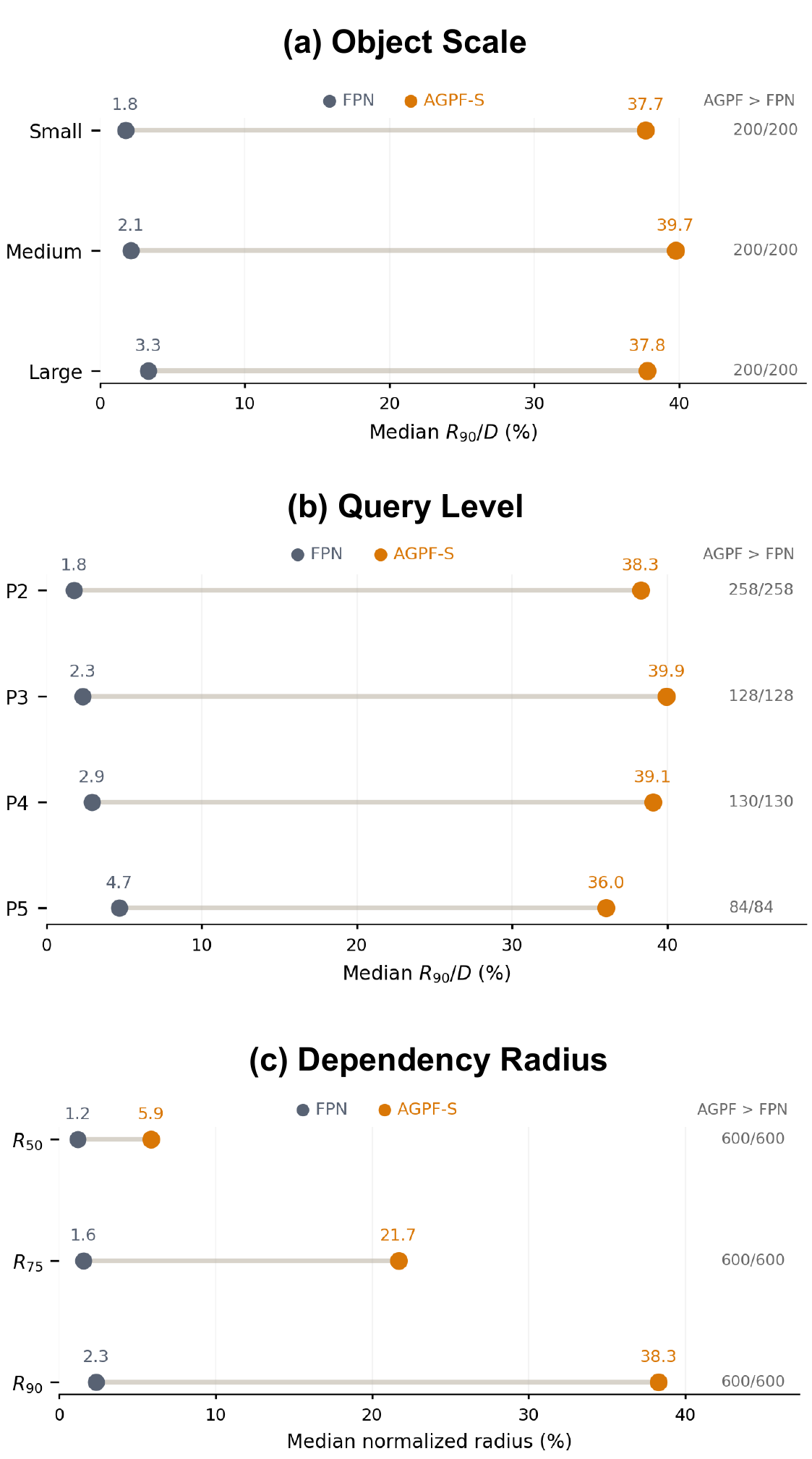}
    \caption{
        Dependency extent across measurement settings.
        Median $R_{90}/D$ is grouped by (a) object scale and
        (b) assigned query level.
        (c) compares radii enclosing $50\%$, $75\%$, and
        $90\%$ of gradient mass.
        All radii are normalized by the reference-grid
        diagonal $D$.
        Counts report paired queries with a larger radius
        for AGPF-S.
    }
    \label{fig:app_erf_robustness}
\end{figure}

\begin{figure*}[t]
    \centering
    \includegraphics[width=\textwidth]
    {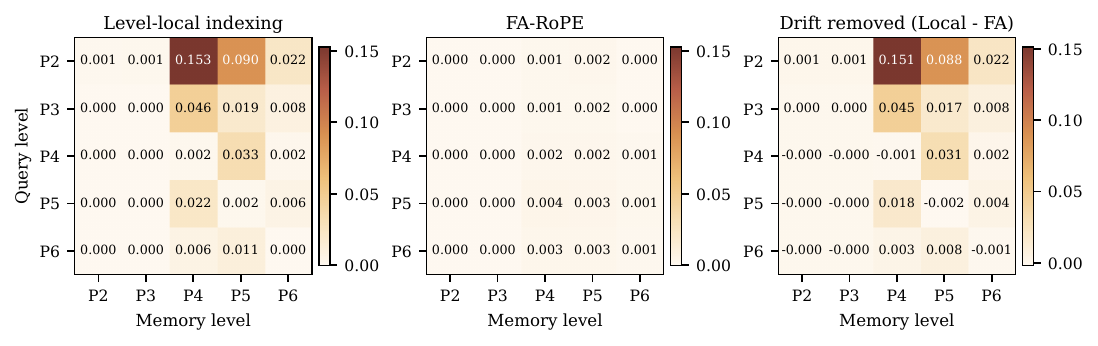}
    \caption{
        Attention drift for each query--memory level pair
        under a $128$-pixel input translation.
        The first two panels compare level-local coordinates
        and FA-RoPE using the same extracted features.
        The right panel shows their difference
        (level-local minus FA-RoPE).
        Positive values indicate reduced drift with FA-RoPE.
    }
    \label{fig:panel_c_supp_pair_drift_shift128}
\end{figure*}

AGPF-S has a larger median $R_{90}/D$ in every displayed
object-size and query-level group.
The difference remains when the enclosed gradient mass is
reduced from $90\%$ to $75\%$ or $50\%$.
The wider spatial dependence is therefore not confined to one
object-size group, query level, or dependency-mass threshold.

\subsection{Per-object paired differences}

Figure~\ref{fig:paired_r90_delta} reports
$\Delta_q=(R_{90,q}^{\mathrm{AGPF}}-R_{90,q}^{\mathrm{FPN}})/D$
for each sampled object.
All $600$ differences are positive.
The median increases are $35.9$, $37.6$, and $33.8$ percentage
points for small, medium, and large objects, respectively.
The larger aggregate extent is thus shared across the sampled
queries rather than driven by a few extreme cases.

\section{Translation Controls for FA-RoPE}
\label{app:positional_consistency}

\subsection{Drift between query and memory levels}

Figure~\ref{fig:panel_c_supp_pair_drift_shift128} breaks down
attention drift by query and memory level under a $128$-pixel
input translation.
Query strides are $[4,8,16,32,64]$, while pooling gives memory
strides of $[16,16,16,32,64]$.
Large reductions occur in several cross-level interactions.
For $P_{2,Q}\!\rightarrow P_{4,K}$, normalized drift falls
from $0.1526$ to $0.00146$, a reduction of approximately $99.0\%$.
FA-RoPE substantially reduces these large discrepancies,
although attention need not be exactly preserved when the
translated image also produces different feature content.

\subsection{Translation with fixed feature content}
\label{app:fixed_content_translation}

\begin{figure}[H]
    \centering
    \includegraphics[width=.94\linewidth]
    {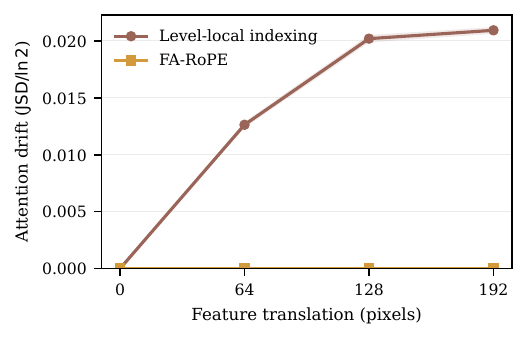}
    \caption{
        Attention drift when extracted pyramid features are
        translated directly.
        Corresponding pre-RoPE queries and keys are held fixed,
        isolating the effect of positional coordinates.
    }
    \label{fig:panel_c_supp_controlled_feature_shift}
\end{figure}

We isolate the positional contribution by translating the
extracted pyramid features directly.
Corresponding pre-RoPE query and key features are identical
before and after translation.
FA-RoPE preserves the measured attention distributions up to
numerical precision
(Fig.~\ref{fig:panel_c_supp_controlled_feature_shift}).
Level-local coordinates instead produce normalized drift of
$0.0126$, $0.0202$, and $0.0209$ for translations of
$64$, $128$, and $192$ pixels, respectively.
Coordinate choice can therefore change cross-resolution
attention even when feature content is unchanged; the shared
spatial reference removes this source of drift in the controlled test.

\section{Enhanced Detector Recipes}
\label{app:system_recipes}

\begin{table}[H]
    \centering
    \small
    \setlength{\tabcolsep}{7pt}
    \setlength{\arrayrulewidth}{0.45pt}
    \renewcommand{\arraystretch}{1.08}
    \begin{tabular}{l|l}
        config                   & value                                            \\
        \specialrule{0.70pt}{0pt}{1pt}
        optimizer                & AdamW                                            \\
        base learning rate       & 2e-4                                             \\
        weight decay             & 0.05                                             \\
        batch size               & 16 (2/GPU, 8 GPUs)                               \\
        warmup iterations        & 1000                                             \\
        training epochs          & 36                                               \\
        learning rate milestones & 27, 33 ($\times 0.1$)                            \\
        augmentation             & multi-scale~\cite{carion2020end}                 \\
        box head                 & 4conv1fc (SyncBN)~\cite{wu2018group}             \\
        box regression loss      & GIoU~\cite{Rezatofighi_2019_CVPR} ($\lambda=10$) \\
        mask loss                & CE + Dice~\cite{milletari2016v} (1:2)            \\
        AGPF blocks              & 6 (L)                                            \\
    \end{tabular}
    \caption{Enhanced training recipe for Cascade Mask R-CNN and HTC.}
    \label{tab:agpf_system_recipes_1}
\end{table}

\begin{table}[H]
    \centering
    \small
    \setlength{\tabcolsep}{7pt}
    \setlength{\arrayrulewidth}{0.45pt}
    \renewcommand{\arraystretch}{1.08}
    \begin{tabular}{l|l}
        config                   & value                           \\
        \specialrule{0.70pt}{0pt}{1pt}
        optimizer                & AdamW                           \\
        base learning rate       & 2e-4                            \\
        weight decay             & 0.05                            \\
        batch size               & 16  (2/GPU, 8 GPUs)             \\
        warmup iterations        & 1000                            \\
        training epochs          & 50                              \\
        learning rate milestones & 40, 48 ($\times 0.1$)           \\
        augmentation             & LSJ~\cite{ghiasi2021simple}     \\
        box head                 & 4conv1fc (SyncBN)               \\
        box regression loss      & GIoU ($\lambda=10$)             \\
        mask loss                & CE + Dice (1:2)                 \\
        AGPF blocks              & 6 (L)                           \\
        RoI NMS                  & Soft-NMS~\cite{Bodla_2017_ICCV} \\
    \end{tabular}
    \caption{Enhanced training and inference recipe for HTC++.}
    \label{tab:agpf_system_recipes_2}
\end{table}

Tables~\ref{tab:agpf_system_recipes_1} and
\ref{tab:agpf_system_recipes_2} summarize the enhanced
detector recipes used for the complete-system comparisons
in Table~\ref{tab:modern_method_comparison}.
Cascade Mask R-CNN and HTC follow a 36-epoch multi-scale
training schedule, while HTC++ adopts a 50-epoch schedule
with large-scale jittering, Copy-Paste augmentation, and
Soft-NMS. No test-time augmentation (TTA) is used.
These enhanced recipes are exclusive to the complete-system
comparisons; all controlled FPN--AGPF experiments retain
the original detector heads, losses, and training settings,
with only the neck replaced.

\section{Additional Qualitative Comparisons}
\label{app:qualitative_gallery}
\begin{figure*}[t]
    \centering
    \includegraphics[width=\linewidth]
    {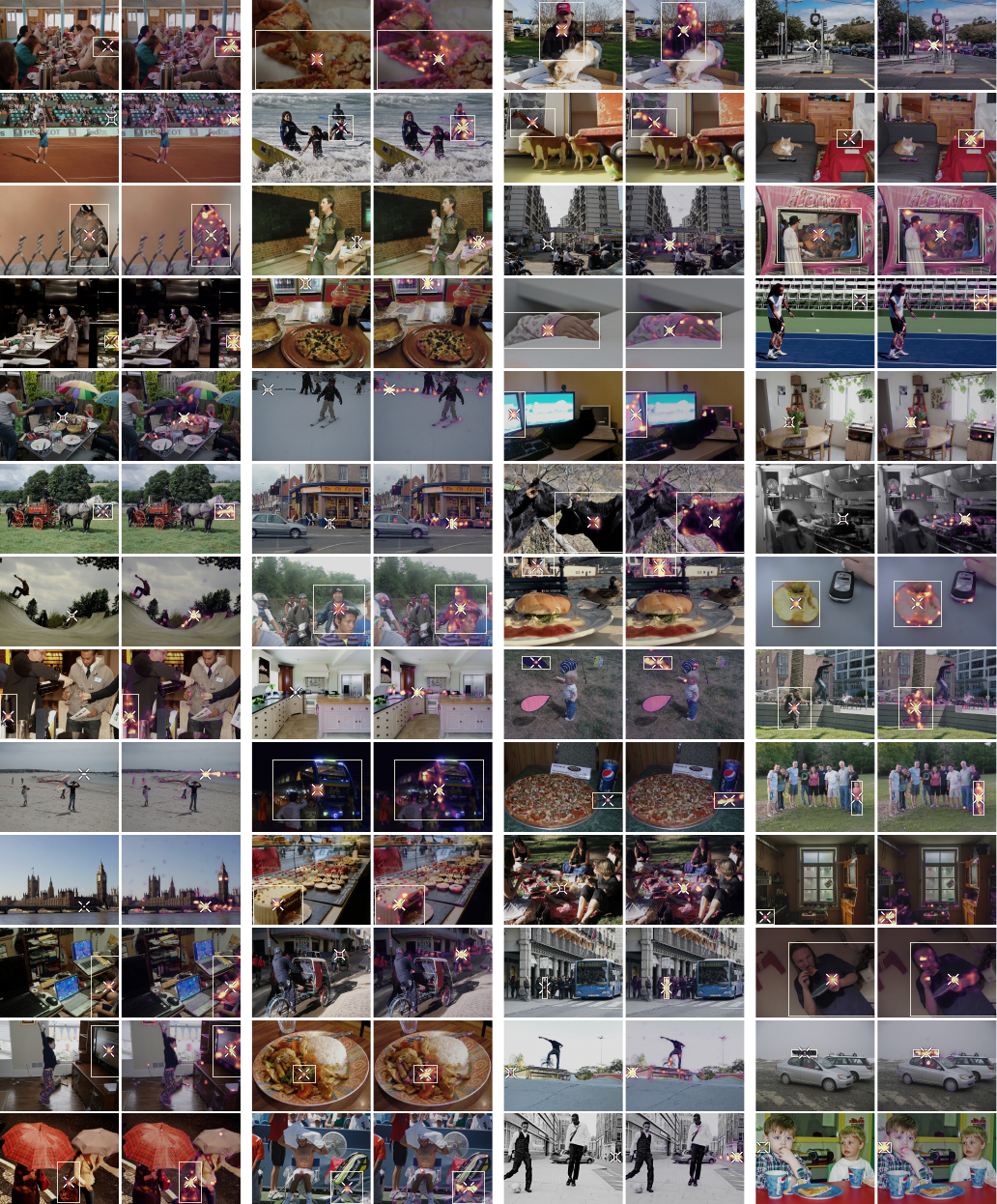}
    \caption{Qualitative spatial dependency maps for matched object-centered queries. Each pair shows FPN (left) and AGPF-S (right).}
    \label{fig:add_qualitative_gallery}
\end{figure*}

Figure~\ref{fig:add_qualitative_gallery} presents additional qualitative comparisons of gradient-based spatial dependency between FPN and AGPF-S on COCO \texttt{val2017}.
For each matched object-centered query, we visualize the sensitivity of its output feature to the input feature pyramid, following the procedure described in Sec.~\ref{sec:approach}.
The white boxes indicate target objects, and the crosses mark query locations.
FPN typically exhibits highly localized responses around the query, whereas AGPF-S maintains local sensitivity while extending its spatial dependence to more distant regions.
These examples complement our quantitative analysis, further illustrating how AGPF enables individual output locations to incorporate information from a broader spatial context.

\end{document}